\documentclass[11pt]{article}
\usepackage{fancyhdr}
\usepackage{titlesec}
\usepackage{titling}

\usepackage{wrapfig}
\usepackage{amsmath}
\usepackage{amsfonts}
\usepackage{amssymb}
\usepackage{amsthm}
\usepackage{epsfig}
\usepackage{natbib}
\usepackage[utf8]{inputenc} 
\usepackage[T1]{fontenc}    
\usepackage{hyperref}       
\usepackage{url}            
\usepackage{booktabs}       
\usepackage[leftcaption]{sidecap}
\usepackage{nicefrac}       
\usepackage{microtype}      
\usepackage{graphicx}
\usepackage{paralist}
\usepackage{enumitem}
\usepackage{xcolor}
\usepackage{comment}
\usepackage{mathtools}
\usepackage{bbm}
\usepackage[linesnumbered,ruled]{algorithm2e}
\usepackage{algorithmic}
\usepackage{thm-restate}
\usepackage{capt-of}
\usepackage{authblk}
\usepackage[symbol,flushmargin]{footmisc}
\usepackage{dblfloatfix}
\usepackage[font={footnotesize}]{caption}
\usepackage{subcaption}
\usepackage{float}

\usepackage[makeroom]{cancel}

\usepackage{mathtools,stackengine}
\stackMath
\newcommand{\stackEq}[1]{%
  \setbox0=\hbox{${}\mathrel{\stackon[-1pt]{=}{\scriptstyle\text{#1\strut}}}{}$}
  \xdef\tmpwd{\dimexpr\the\wd0\relax}
  \kern.5\tmpwd\mathclap{\box0}&\kern.5\tmpwd
}

\def\secref#1{section~\ref{#1}}

\def\eqref#1{equation~\ref{#1}}

\def\1{\bm{1}}

\def\rva{{\mathbf{a}}}

\def\rvc{{\mathbf{c}}}

\def\rvm{{\mathbf{m}}}

\def\rvo{{\mathbf{o}}}

\def\rvs{{\mathbf{s}}}

\def\rvx{{\mathbf{x}}}
\def\rvy{{\mathbf{y}}}

\DeclareMathAlphabet{\mathsfit}{\encodingdefault}{\sfdefault}{m}{sl}
\SetMathAlphabet{\mathsfit}{bold}{\encodingdefault}{\sfdefault}{bx}{n}

\newcommand{\E}{\mathbb{E}}

\DeclareMathOperator*{\argmax}{arg\,max}

\newcommand{\mc}[1]{{\mathcal{#1}}}

\newcommand{\RR}{\mathbb{R}}

\renewcommand{\S}{\mathcal{S}}
\newcommand{\A}{\mathcal{A}}

\newcommand{\X}{\mathcal{X}}
\newcommand{\Y}{\mathcal{Y}}

\newcommand{\LL}{\mathcal{L}}
\renewcommand{\O}{\mathcal{O}}

\newcommand{\N}{\mathcal{N}}

\makeatletter
\renewcommand\AB@affilsepx{}
\makeatother

\begin{document}

\title{{\LARGE\bfseries Active World Model Learning with Progress Curiosity}}

\date{}

\author[1]{Kuno Kim}
\author[1]{Megumi Sano}
\author[2]{Julian De Freitas}
\author[3*]{Nick Haber}
\author[1, 4, 5*]{Daniel Yamins}

\affil[1]{Department of Computer Science, Stanford University \quad} 
\affil[2]{Department of Psychology, Harvard University \newline}
\affil[3]{Stanford Graduate School of Education \quad}
\affil[4]{Department of Psychology, Stanford University \newline}
\affil[5]{Wu Tsai Neuroscience Institute, Stanford University \quad}
\affil[*]{Equal contribution.}

\maketitle
\vspace{-2cm}
 
\thispagestyle{empty}

\begin{abstract}
World models are self-supervised predictive models of how the world evolves. Humans learn world models by curiously exploring their environment, in the process acquiring compact abstractions of high bandwidth sensory inputs, the ability to plan across long temporal horizons, and an understanding of the behavioral patterns of other agents. In this work, we study how to design such a curiosity-driven Active World Model Learning (AWML) system. To do so, we construct a curious agent building world models while visually exploring a 3D physical environment rich with distillations of representative real-world agents. We propose an AWML system driven by $\gamma$-Progress: a scalable and effective learning progress-based curiosity signal. We show that $\gamma$-Progress naturally gives rise to an exploration policy that directs attention to complex but learnable dynamics in a balanced manner, thus overcoming the ``white noise problem''. As a result, our $\gamma$-Progress-driven controller achieves significantly higher AWML performance than baseline controllers equipped with state-of-the-art exploration strategies such as Random Network Distillation and Model Disagreement. 
\end{abstract}\footnote[0]{\noindent A slightly abridged version of this paper was published in the \textit{Proceedings of the 37th International Conference on Machine Learning} (ICML 2020).}
\section{Introduction} 
\label{sec:introduction}

Imagine yourself as an infant in your parent's arms, sitting on a playground bench. You are surrounded by a variety  of potentially interesting stimuli, from the constantly whirring merry-go-round, to the wildly rustling leaves, to your parent's smiling and cooing face.
After briefly staring at the motionless ball, you grow bored. 
You consider the merry-go-round a bit more seriously, but its periodic motion is ultimately too predictable to keep your attention long. 
The leaves are quite entertaining, but after watching their random motions for a while, your gaze lands on your parent. Here you find something really interesting: you can anticipate, elicit, and guide your parents' changes in expression as you both engage in a game of peekaboo. 
Though just an infant, you have efficiently explored and interacted with the environment, in the process gaining strong intuitions about how different things in your world will behave. 

The infant appears to have learned a powerful \emph{world model} --- a predictor of how the world evolves over time, due both to external physical dynamics and to the infant's actions.
Such world models help enable humans to plan across long temporal horizons and to anticipate the behavioral patterns of other agents. 
They also may play an important role in the self-supervised learning of the high-bandwidth sensory systems that produce compact perceptual abstractions underlying cognition and decision making. 
Devising algorithms that can efficiently construct such world models is an important goal for the next generation of socially-integrated AI and robotic systems.

A key challenge in world model learning is that real-world environments contain a diverse range of dynamics, generated by a multiplicity of objects and other agents, with varying levels of learnability. 
The inanimate ball and periodic merry-go-round display dynamics that are easy to learn. 
On the other hand, stimuli such as falling leaves exhibit unlearnable noise-like dynamics. 
Lying in a ``sweet spot'' on the learnability spectrum are animate agents that generate interesting and complex yet rule-driven dynamics, e.g. your parent's expressions and play offerings. 
Balancing attention to maximize learning progress amidst the blooming and buzzing sea of stimuli is a substantial challenge. Particularly difficult is the \emph{white noise problem} ~\citep{schmidhuber_formaltheoryoffun, burda2018exploration, pathak2019disagreement}, i.e. perseverating on unlearnable stimuli rather than pursuing learnable dynamics.
Thus, it is a natural hypothesis that behind the infant's ability to learn powerful world models must be an equally powerful \emph{active learning} algorithm that directs its attention to maximize learning progress. 

In this work, we formalize and study Active World Model Learning (AWML) -- the problem of determining a directed exploration policy that enables efficient construction of better world models in agent-rich contexts. To do so, we construct a progress-driven curious neural agent performing AWML in a custom-built 3D virtual world environment. Specifically, our contributions are as follows:
\begin{enumerate}
    \item We construct a 3D virtual environment rich with agents displaying a wide spectrum of realistic stimuli behavior types with varying levels of learnability, such as static, periodic, noise, peekaboo, chasing, and mimicry.
    \item We formalize AWML within a general reinforcement learning framework that encompasses curiosity-driven exploration and traditional active learning. 
    \item We propose an AWML system driven by $\gamma$-Progress: a novel and scalable learning progress-based curiosity signal. We show that $\gamma$-Progress gives rise to an exploration policy that 
    overcomes the white noise problem and achieves significantly higher AWML performance than state-of-the-art exploration strategies — including Random Network Distillation (RND) \citep{burda2018exploration} and Model Disagreement \citep{pathak2019disagreement}.
\end{enumerate}

\section{Related Works}
\label{sec:relatedworks}

\subsection{Artificial Intelligence Literature}

\textbf{World Models.}
A natural class of world models involve forward dynamics prediction. Such models can directly predict future video frames~\citep{finn2016unsupervised, wang2018predrnn++, Wu2019ModelIF}, or latent feature representations such as 3D point clouds~\citep{byravan2017se3} or object-centric, graphical representations of scenes~\citep{battaglia2016interaction, chang2016compositional, mrowca2018flexible}. 
Action-conditioned forward-prediction models can be used directly in planning for robotic control tasks~\citep{finn2017deep}, as performance-enhancers for reinforcement learning tasks~\citep{ke2019learning}, or as ``dream'' environment simulations for training policies~\citep{ha2018world}. In our work, we focus on forward dynamics prediction with object-oriented representations.  \\

\noindent \textbf{Active Learning and Curiosity.} 
A key question the agent is faced with is how to choose its actions to efficiently learn the world model. In the classical \emph{active learning} setting~\citep{settles2011_active}, an agent seeks to learn a supervised task with costly labels, judiciously choosing which examples to obtain labels for so as to maximize learning efficiency. 
More recently, active learning has been implicitly generalized to self-supervised reinforcement learning agents~\citep{schmidhuber_formaltheoryoffun, oudeyer2013intrinsically, jaderberg2016reinforcement}. In this line of work, agents typically self-supervise a world model with samples obtained by curiosity-driven exploration. Different approaches to this general idea exist, many of which are essentially different approaches to estimating future \emph{learning progress} --- e.g. determining which actions are likely to lead to the highest world model prediction gain in the future. 
One approach is the use of \emph{novelty} metrics, which measure how much a particular part of the environment has been explored, and direct agents into under-explored parts of state-space. 
Examples include count-based and psuedo-count-based methods~\citep{strehl2008analysis,bellemare2016unifying, ostrovski2017count}, Random Network Distillation (RND)~\citep{burda2018exploration}, and \emph{empowerment} \citep{mohamed2015variational}. 
Novelty-based approaches avoid the difficult world model  progress estimation problem entirely by not depending at all on a specific world model state, and relying on novelty as a (potentially inconsistent) proxy for expected learning progress. 

The simplest idea that takes into account the world model is \emph{adversarial} curiosity, which estimates current world model error and directs agents to take actions estimated to maximize this error~\citep{stadie2015predictive, pathak2017inverse, haber2018learning}. However, adversarial curiosity is especially prone to the \emph{white noise problem}, in which agents are motivated to waste time fruitlessly trying to solve unsolvable world model problems, e.g. predicting the dynamics of random noise. The white noise problem can to some degree be avoided by solving the world-modeling problem in a learned latent feature space in which degeneracies are suppressed~\citep{pathak2017inverse, pathak18largescale}. 

Directly estimating learning progress ~\citep{oudeyer2007intrinsic, oudeyer2013intrinsically} or \emph{information gain}~\citep{houthooft_vime} avoids the white noise problem in a more comprehensive fashion. However, such methods have been limited in scope because they involve calculating quantities that cannot easily be estimated in high-dimensional continuous action spaces.
\emph{Surprisal}~\citep{achiam2017surprise} and model disagreement~\citep{pathak2019disagreement} present computationally-tractable alternatives to information gain, at the cost of the accuracy of the estimation. For comprehensive reviews of intrinsic motivation signal choices, see \citep{aubret2019survey, linke2019adapting}.
In this work, we present a novel method for estimating learning progress that is ``consistent'' with the original prediction gain objective while also scaling to high-dimensional continuous action-spaces.

\subsection{Cognitive Science Literature}
\label{subsec:related:cognitive}

\textbf{Intuitive physics and object-based priors.}  
Humans excel at intuitively predicting object dynamics ~\citep{battaglia2018relational}. A key framework underlying such abilities is object-centric attention allocation. Humans are able to keep track of objects over time, even as they become occluded or leave the visual frame~\citep{piaget_originsintelligence}. In this work, we include object-based attention and object permanence as neural architectural biases.  \\

\noindent \textbf{Curiosity and active learning.} 
Humans interact with the world to learn how it works. Infants actively gather information from their environment by attending to objects in a highly non-random manner \citep{smith2019modeling}, devoting more attention to objects that violate their expectations \citep{stahl2015observing}. They also self-generate learning curricula, preferring stimuli that are complex enough to be interesting but still predictable~\citep{kidd2012goldilocks}. We study active learning by means of attention allocation.  \\

\noindent \textbf{Animate attention.}
From early infancy, humans effectively distinguish between inanimate and animate agents, preferentially paying attention to animate features like faces \citep{maurer2002many}. Even in the absence of such visual features, infants preferentially attend to spatiotemporal kinematics indicative of animacy, such as efficient movement towards  targets~\citep{gergely1995taking} and contingent behavior between agents~\citep{frankenhuis2013infants}. Such kinematic patterns give rise to an irresistable sense of animacy, even when the moving objects are simple shapes~\citep{heider1944experimental}. 
In this work, instead of injecting biases for animate attention, we test whether it emerges naturally, albeit with the right choice of curiosity. \\

\noindent \textbf{Social prediction and theory of mind.}
A more sophisticated ability emerging later in development is understanding other agents' behaviors as consequences of their underlying mental states, aka \emph{theory of mind} \citep{astington1990developing}. Theory of mind allows generating predictions about other agents as a function of their underlying mental states. 
In this work, our model learns to predict what other agents in the environment will do next through the use of a disentangled architecture that leverages the idea of different agents having different underlying internal states.

\begin{figure}[t]
    \centering
    \includegraphics[width=\textwidth]{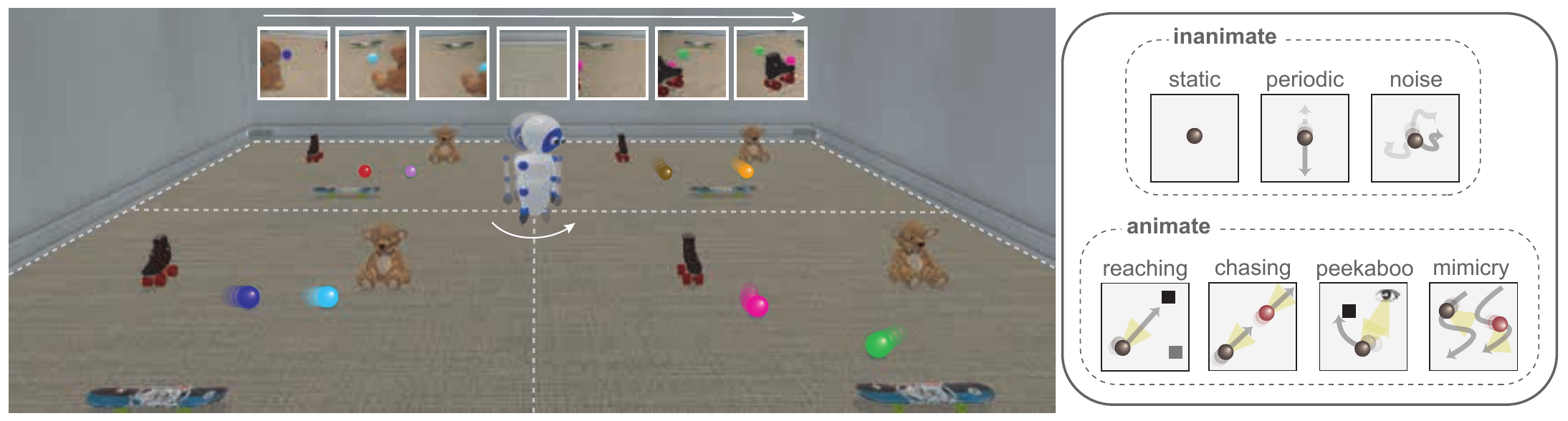} 
    \caption{\textbf{Virtual environment.} Our 3D virtual environment is a distillation of key aspects of real-world environments. The \textit{curious agent} (white robot) is centered in a room, surrounded by various \textit{external agents} (colored spheres) contained in different quadrants, each with dynamics that correspond to a realistic inanimate or animate behavior (right box). The curious agent can rotate to attend to different behaviors as shown by the first-person view images at the top. See \textcolor{blue}{\url{https://bit.ly/31vg7v1}} for videos.} 
    \label{fig:environment}
\end{figure}
    
\section{Multi-Agent Virtual World Environment}
\label{sec:environment}
To faithfully replicate real-world algorithmic challenges, we design our 3D virtual environment to preserve the following key properties of real-world environments: 
\vspace{-5pt} 
\begin{enumerate} 
    \item \textit{Diverse dynamics.} Agents operate under a diverse set of dynamics specified by agent-specific programs. An agent's actions may depend on those of another agent resulting in complex interdependent relationships.
    
    \item \textit{Partial observability.} At no given time do we have full access to the state of every agent in the environment. Rather, our learning is limited by what lies within our field of view. 
    
    \item \textit{Contingency.} How much we learn is contingent on how we, as embodied agents, choose to interact with the environment. 
\end{enumerate} 
\vspace{-5pt}
Concretely, our virtual environment consists of two main components, a \textit{curious agent} and various \textit{external agents}.

The \textit{curious agent}, embodied by an avatar, is fixed at the center of a room (Figure \ref{fig:environment}). Just as a human toddler can control her gaze to visually explore her surroundings, the agent is able to partially observe the environment based on what lies in its field of view (see top of Figure \ref{fig:environment}). The agent can choose from 9 actions: rotate $12^{\circ}, 24^{\circ}, 48^{\circ}$, or $96^{\circ}$, to the left/right, or stay in its current orientation. 

The \textit{external agents} are spherical avatars that each act under a hard-coded policy inspired by real-world inanimate and animate stimuli. An \textit{external agent behavior} consists of either one external agent, e.g reaching, or two interacting ones, e.g chasing. Since external agents are devoid of surface features, the curious agent must learn to attend to different behaviors based on spatiotemporal kinematics alone. We experiment with external agent behaviors (see Figure \ref{fig:environment}, right) including static, periodic, noise, reaching, chasing, peekaboo, and mimicry. The animate behaviors have deterministic and stochastic variants, where the stochastic variant preserves the core dynamics underlying the behavior, albeit with more randomness. See \textcolor{blue}{\url{https://bit.ly/31vg7v1}} for video descriptions of the environment and external agent behaviors. 

We divide the room into four quadrants, each of which contains various auxiliary objects (e.g teddy bear, roller skates, surfboard) and one external agent behavior. The room is designed such that the curious agent can see at most one external agent behavior at any given time. This design is key in ensuring partial observability, such that the agent is faced with the problem of allocating attention between different external agent behaviors in an efficient manner. 
Below, we describe all behaviors in detail. Note that a subset of behaviors (peekaboo, reaching, and chasing) is further sub-divided into deterministic and stochastic varieties. \\

\noindent \textbf{Inanimate behaviors}
    
\newcommand\imgr{0.15}
\newcommand\textr{0.85}

\begin{minipage}[t]{\imgr\textwidth}
    \centering
    \strut\vspace*{-\baselineskip}\newline
    \includegraphics[width=\textwidth]{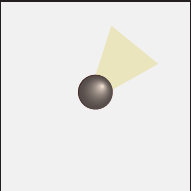}
\end{minipage}%
\hspace{5pt}
\begin{minipage}[t]{\textr\textwidth}
    \textbf{\textit{Static}} Inspired by stationary objects such as couches, lampposts, and fire hydrants, the \textit{static agent} remains at its starting location and stays immobile.
\end{minipage}

\begin{minipage}[t]{\imgr\textwidth}
    \centering
    \strut\vspace*{-\baselineskip}\newline
    \includegraphics[width=\textwidth]{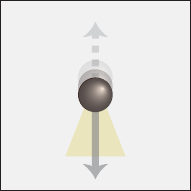}
\end{minipage}%
\hspace{5pt}
\begin{minipage}[t]{\textr\textwidth}
    \textbf{\textit{Periodic}} Inspired by objects exhibiting periodic motion such as fans, flashing lights, and clocks, the \textit{periodic agent} regularly moves back and forth between two specified locations in its quadrant.
\end{minipage}

\begin{minipage}[t]{\imgr\textwidth}
    \centering
    \strut\vspace*{-\baselineskip}\newline
    \includegraphics[width=\textwidth]{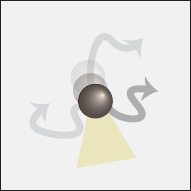}
\end{minipage}%
\hspace{5pt}
\begin{minipage}[t]{\textr\textwidth}
    \textbf{\textit{Noise}} Inspired by random motion in wind, water, and other inanimate elements, the \textit{noise agent} randomly samples a new direction and moves in that direction with a fixed step size while remaining within the boundaries of its quadrant. 
\end{minipage}
\vspace{10pt}

\noindent \textbf{Animate Behaviors}

\begin{minipage}[t]{\imgr\textwidth}
    \centering
    \strut\vspace*{-\baselineskip}\newline
    \includegraphics[width=\textwidth]{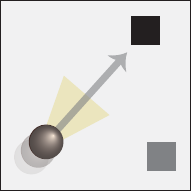}
\end{minipage}%
\hspace{5pt}
\begin{minipage}[t]{\textr\textwidth}
    \textbf{\textit{Reaching (deterministic)}} We often exhibit goal-oriented behavior by interacting with objects. The \textit{reacher agent} approaches each auxiliary object in its quadrant sequentially, such that object positions fully determine its trajectory. Objects periodically shift locations such that predicting agent behavior at any given time requires knowing the current object positions.
\end{minipage}
    
\begin{minipage}[t]{\imgr\textwidth}
    \centering
    \strut\vspace*{-\baselineskip}\newline
    \includegraphics[width=\textwidth]{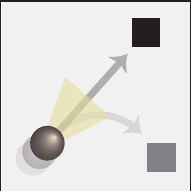}
\end{minipage}%
\hspace{5pt}
\begin{minipage}[t]{\textr\textwidth}
    \textbf{\textit{Reaching (stochastic)}} The order in which the reacher agent visits the objects is stochastic  (uniform sampling from the three possible objects). However, once the reacher agent starts moving towards an object, its trajectory for the next few time steps, before it chooses a different object to move to, is predictable. 
\end{minipage}  
    
\begin{minipage}[t]{\imgr\textwidth}
    \centering
    \strut\vspace*{-\baselineskip}\newline
    \includegraphics[width=\textwidth]{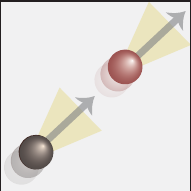}
\end{minipage}%
\hspace{5pt}
\begin{minipage}[t]{\textr\textwidth}
    \textbf{\textit{Chasing (deterministic)}} We often act contingently on the actions of other agents, which in turn depend on our own. In chasing, a \textit{chaser agent} chases a \textit{runner agent}. If the runner is too close to quadrant bounds, it then escapes to one of a few escape locations away from the chaser but within the quadrant. Thus, the chaser's position affects the runner's trajectory and vice versa. 
\end{minipage}  

\begin{minipage}[t]{\imgr\textwidth}
    \centering
    \strut\vspace*{-\baselineskip}\newline
    \includegraphics[width=\textwidth]{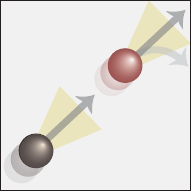}
\end{minipage}%
\hspace{5pt}
\begin{minipage}[t]{\textr\textwidth}
    \textbf{\textit{Chasing (stochastic)}} When the runner agent is too close to the quadrant bounds, it escapes by picking any random location away from the chaser and within the bounds of the quadrant. 
\end{minipage}  

\begin{minipage}[t]{\imgr\textwidth}
    \centering
    \strut\vspace*{-\baselineskip}\newline
    \includegraphics[width=\textwidth]{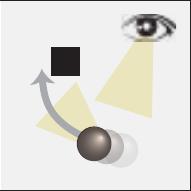}
\end{minipage}%
\hspace{5pt}
\begin{minipage}[t]{\textr\textwidth}
\textbf{\textit{Peekaboo (deterministic)}} The \textit{peekaboo agent} acts contingently on the curious agent. If the curious agent stares at it, it hides behind an auxiliary object such as a doll. If the curious agent continues to stare, it starts \textit{peeking} out by moving to a close fixed location. Once the curious agent looks away, it stops hiding, returning to an exposed location.  
\end{minipage}  

\begin{minipage}[t]{\imgr\textwidth}
    \centering
    \strut\vspace*{-\baselineskip}\newline
    \includegraphics[width=\textwidth]{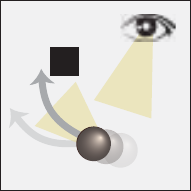}
\end{minipage}%
\hspace{5pt}
\begin{minipage}[t]{\textr\textwidth}
    \textbf{\textit{Peekaboo (stochastic)}}  There are multiple peeking locations near the hiding object that the peekaboo agent can visit randomly during its peeking behavior. 
\end{minipage}  

\begin{minipage}[t]{\imgr\textwidth}
    \centering
    \strut\vspace*{-\baselineskip}\newline
    \includegraphics[width=\textwidth]{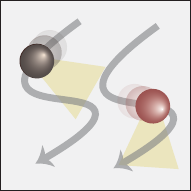}
\end{minipage}%
\hspace{5pt}
\begin{minipage}[t]{\textr\textwidth}
    \textbf{\textit{Mimicry (deterministic)}} From an early age, we learn by imitating others.
    Mimicry consists of an \textit{actor agent} (red) and an \textit{imitator agent} (gray), each staying in one half of the quadrant to avoid collisions. The actor acts identically to the random agent, while the imitator mirrors the actor's trajectory with a delay, such that the past trajectory of the actor fully determines the future trajectory of the imitator. 
\end{minipage}  

\begin{minipage}[t]{\imgr\textwidth}
    \centering
    \strut\vspace*{-\baselineskip}\newline
    \includegraphics[width=\textwidth]{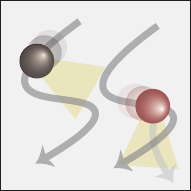}
\end{minipage}%
\hspace{5pt}
\begin{minipage}[t]{\textr\textwidth}
    \textbf{\textit{Mimicry (stochastic)}} The imitator agent is imperfect and produces a noisy reproduction of the actor agent's trajectory. 
\end{minipage}

\section{Theory}
\label{sec:theory}
In this section we formalize Active World Model Learning (AWML) as a Reinforcement Learning (RL) problem that is a specific form of active learning. We then discuss a number of curiosity signals that can be used to drive AWML, and introduce $\gamma$-Progress, a scalable progress-based measure with several algorithmic and computational advantages over previous signals.  

\subsection{Active World Model Learning}

We formalize an agent in environment as the tuple $\mc{E} := (\S, \A, P, P_{0})$. $\S$ denotes the set of states the agent and environment can be in --- in the virtual world environment described in \secref{sec:environment}, $\S$ captures the gaze direction of the curious agent, the positions and type of external objects, and the positions and internal states of the external agents\footnote{Our virtual world environment is {\em partially observable} and hence requires the additional specification of $\O$, the set of observations, and $Q = Q(\rvo | \rvs, \rva)$, the set of conditional observation probabilities. For the sake of simplicity, we suppress this complication in the main text and point out where it is salient in a series of footnotes.}.  $\A$ represents the set of actions the agent can take, and are constrained by the physical avatar of the agent --- in the virtual world, the choice of how far and where to turn its gaze.  Transition dynamics are given by the function $P: \S \times \A \rightarrow \Omega(\S)$, where $\Omega(\S)$ is the set of probability measures on $\S$ (allowing for stochastic environment dynamics). In the case of our virtual world, $P$ captures both the effect of the gaze actions of the agent (e.g. changes in which part of the scene is being observed), as well the dynamics of each of the external agents.  The function $P_{0}: \S \rightarrow [0, 1]$ describes the probability distribution of initial conditions of states. 


In this environment, the agent's overall goal is to learn a target function $\omega$ with as few data samples as possible.  In general, $\omega$ can be any predictor on finite-horizon state-action trajectories sampled from the environment. That is, $\omega: \X \rightarrow \Omega(\Y),$ where $\X := \S^{i_s} \times \A^{i_a}$ and $\Y := \Omega(\S^{o_s} \times \A^{o_a})$ represent sets of fixed-length observation-action sequences. (The non-negative integers $i_s, i_a, o_s,$ and $o_a$ are the input and output state and action horizons, respectively.)
In this work, we work with forward prediction, i.e. the situation where $\X = \S \times \A, \Y = \S$, and $\omega = P$, but a variety of other potentially useful targets, such as inverse prediction, can also be formulated by appropriate choice of $\X, \Y$ and $\omega$.\footnote{Actually, in partial observable case such as ours, the agent predicts observations from observations rather than raw states from raw states. Observations can contain additional information, such as the direction an external agent is moving, that is relevant to predicting future observations. However, they are also also typically quite incomplete, e.g. if an external agent is invisible until the observation to be predicted. This partial observability leads to what can be thought of as additional white noise, or {\em degeneracy} in the world model problem~\citep{haber2018learning}.} 

The agent seeks to estimate a parameterized model $\omega_{\theta}$ of $\omega$ (e.g $\theta$ are parameters deep neural network; see \secref{sec:methods} below). We henceforth refer to $\omega_{\theta}$ as the world model.
To measure its error during world model optimization, the agent is equipped with a loss function $\LL: (x, f, g) \mapsto \RR$ such that for any $x \in \X$ and any functions $f,g: \X \rightarrow \Omega(\Y)$, $\LL(x, f, g)$ achieves its minimum whenever $f(x) = g(x)$.  
A measure $\mu$ over $\X$ representing a validation data distribution is also specified, so that the agent's learning goal is to minimize $\LL_{\mu}(\theta) := \E_{\mu}[\LL(\theta)] = \int_{\X} \LL(x, \omega(x), \omega_{\theta}(x)) \mu(x) dx$. 

The agent learns the world model from data gathering by acting in the environment. We formally define Active World Model Learning as a Markov Decision Process (MDP) $\mc{M} := (\bar{\S}, \bar{\A}, \bar{P}, \bar{P}_{0}, r)$ with state and action spaces $\bar{\S}, \bar{\A}$, dynamics and initial conditions $\bar{P}, \bar{P}_0$, and reinforcement reward function $r$. Because intrinsically-motivated policies (such as progress curiosity) will critically depend on states of the agent's world model, $\mc{M}$ is an augmentation of the environment $\mc{E}$ that is constructed by adding the data-collection and model parameter history of the agent itself. 
 
Specifically, the augmented state space $\bar{\S} := \S \times \mc{H} \times \Theta$, so that $\bar{\rvs} \in \bar{\S}$ has the form $\bar{\rvs} = (\rvs, H, \theta)$. $\rvs \in \S$ is an environment state, $H = (\rvs_{0}, \rva_{0}, \rvs_{1}, \rva_{1} \ldots) \in \mc{H}$ is the history of environment state-actions visited so far, and $\theta \in \Theta$ is the current model parameters. The action space $\bar{\A} := \A$ is simply the same set of actions available to the agent in the environment\footnote{In the partial observability case, the action choice determines not only the state transition but also what is observable each timestep, and hence the agent should keep the interesting in view. The MDP becomes a POMDP, where we assume that the agent has full access to its internal state and history, so augmented observations $\bar{o} \in \bar{\mathcal{O}} = \mathcal{O} \times \mc{H} \times \Theta$ has the form $\bar{o} = (o, H, \theta$), where $o \in O$ (augmented conditional observation probabilities $\bar{Q}$ are similarly derived from $Q$).}. The dynamics are described by $\bar{P}: \bar{\S} \times \A \rightarrow \Omega(\bar{\S})$, which step $\rvs$ according to the environment dynamics $P$, augment the history with new data, and updates the world model $\omega_{\theta}$ on the augmented history. Formally this is described by the sampling procedure: 
\begin{align*}
(\rvs', H', \theta') \sim \bar{P}(\cdot | \bar{\rvs} = (\rvs, H, \theta), \rva) ~~ \text{where } \rvs' \sim P(\rvs, \rva),~H' = H \cup \{\rva, \rvs'\},~\theta' \sim P_{\ell}(H', \theta) 
\end{align*}
where $P_{\ell}: H \times \Theta \rightarrow \Omega(\Theta)$ is a (stochastic) update rule for the world model parameters, e.g. a (stochastic) learning algorithm which updates the parameters on the history of data.
The initial conditions $\bar{P}_{0}(\bar{\rvs} = (\rvs, H, \theta)) = P_{0}(\rvs)\mathbbm{1}(H = \{\})q(\theta)$ is the augmented initial-distribution where $\mathbbm{1}$ is the indicator function and $q(\theta)$ is a prior distribution over the model parameters. 

The function $r$ encodes the learning objective of the agent as an RL reward.  A policy is a map $\pi: \S \rightarrow \Omega(\A)$ from states to action distributions.  In general, the infinite-horizon RL problem is to find an optimal policy $\pi^* = \argmax_{\pi} J(\pi)$, where $J(\pi) = \E_{\pi}[\sum_{t = 0}^{\infty} \beta^{t} r_{t}]$ and $0 \leq \beta < 1$ is a discount factor.  
The goal of AWML in specific is to make effective data-collection decisions to minimize world model loss. 
This can in theory be accomplished by taking the reward function of AWML to be 
\begin{align}
\label{eq:base_cost}
r(\bar{\rvs}, \rva, \bar{\rvs}') = - \LL_{\mu}(\theta'),
\end{align}
where $\bar{\rvs} = (\rvs, H, \theta), \bar{\rvs}' = (\rvs', H', \theta')$ and $\theta' = P_{\ell}(H \cup \{\rva, \rvs'\}, \theta)$ is the updated model parameters after collecting new data $\{\rva, \rvs'\}$. 

It is useful to note that, given that the definition of total reward $J$ is a telescoping geometric sum, optimizing for eq \ref{eq:base_cost} is essentially equivalent to optimizing for the reward function:
\begin{align}
r(\bar{\rvs}, \rva, \bar{\rvs}') = \LL_{\mu}(\theta) - \LL_{\mu}(\theta').
\label{eq:cost}
\end{align}
Thus we can see that, $r(\bar{\rvs}, \rva, \bar{\rvs}')$ essentially measures the reduction in world model loss as a result of obtaining new data $\{\rva, \rvs'\}$, i.e the \emph{prediction gain}.

By appropriately constructing $\mc{M}$, different variants of traditional active learning can be recovered as AWML problems.
For example, Query Synthesis Active Learning~\citep{settles2011_active} is obtained by taking $\S = \Y, \A = \X$, and $P(\cdot |  \cdot, \rva=\rvx) = \omega(\rvx)$. In words, the agent proposes a synthetic data query $\rva$ and the oracle $P$ provides a label $\rvs'$. Other traditional active learning tasks can also be derived, including pool-based and stream active learning (see Appendix \ref{sec:app:gal_cal_connection} for details).  

However, there are several complications making it challenging to use eq.~\ref{eq:cost} directly.  
First, $\mu$ can be a rather diffuse distribution which makes it intractable to compute eq.~\ref{eq:cost} at every environment step.
This is especially problematic in the types of environments of interest here and in other recent works on curiosity-driven learning, relative to the more constrained situations of traditional active learning. 
Secondly, in cases in which an agent explores an unknown environment, $\mu$ is not even known prior to interacting with the environment. 
These bottlenecks necessitate an efficiently-computable heuristic reward function that will typically promote the same learning goal of eq.~\ref{eq:cost} --- constructing a learning dataset that minimizes the loss $\LL_{\mu}$ --- while being independent of any particular choice of $\mu$.  The literature on algorithmic curiosity has explored many variants of such heuristic ``curiosity signals'', which achieve consistency with the learning goals of eq.~\ref{eq:cost} with varying degrees of accuracy and efficiency. A spectrum of such ideas, including our novel proposal ($\gamma$-Progress), are described in the next section. 

\subsection{Curiosity Signals}
\label{sec:app:curiosity_connection}
We now motivate $\gamma$-Progress by outlining the limitations of previously proposed curiosity signals and highlighting the computational and algorithmic advantages of our method. \\

\noindent \textbf{Information Gain} \citep{houthooft_vime, linke2019adapting} based methods seek to minimize uncertainty in the Bayesian posterior distribution over model parameters: 
\begin{equation}
r(\bar{\rvs}, \rva, \bar{\rvs}') = D_{\text{KL}}(p(\theta') || p(\theta))
\end{equation}
where $p(\theta') = p(\theta | H \cup \{\rva, \bar{\rvs}'\})$ and $p(\theta) = p(\theta | H)$. Note that, information gain is a lower bound to the prediction gain under weak assumptions \citep{bellemare2016unifying}. If the posterior has a simple form such as Laplace or Gaussian, information gain can be estimated by weight change $|\theta' - \theta|$ \citep{linke2019adapting}, and otherwise one may resort to learning a variational approximation $q$ to approximate the information gain with $D_{\text{KL}}(q(\theta') || q(\theta))$ \citep{houthooft_vime}. The former weight change methods require a model after every step in the environment and is thus impractical in many settings where world model updates are expensive, e.g. backpropagation through deep neural nets. The latter family of variational methods require maintenance of a parameter distribution and an interlaced evidence lower bound optimization and are thus impractical to use with modern deep nets \citep{achiam2017surprise}. \\

\noindent \textbf{Adversarial} \citep{stadie2015predictive, pathak2017inverse, haber2018learning} curiosity assumes prediction gain is proportional to the current world model loss, which, for forward prediction AWML with negative log likelihood loss, is
\begin{equation}
r(\bar{\rvs}, \rva, \bar{\rvs}') = -\log{\omega_{\theta}(\rvs' | \rvs, \rva)}.
\label{eq:adversarial}
\end{equation}
This assumption holds when the target function $\omega$ is learnable by the model class $\Theta$ and the learning algorithm $P_{\ell}$ makes monotonic improvement without the need for curriculum learning. However, adversarial reward is perpetually high when the target is unlearnable by the model class, e.g. deterministic model $\omega_{\theta}$ cannot match stochastic target $\omega$ on inputs $\rvx$ for which $\omega(\rvx)$ is not a Dirac-delta function. As a result, the curious agent suffers from the white noise problem \citep{schmidhuber_formaltheoryoffun}, i.e it endlessly fixates on unlearnable stimuli. \\


\noindent \textbf{Disagreement} \citep{pathak2019disagreement} assumes future world model loss reduction is proportional to the prediction variance of an ensemble of $N$ world models $\{P_{\theta_{j}}\}_{j = 1}^{N}$.
\begin{equation}
r(\bar{\rvs}, \rva, \bar{\rvs}') = \text{Var}(\{\omega_{\theta_{j}}(\rvs' | \rvs, \rva)\}_{j = 1}^{N})
\label{eq:disagreement}
\end{equation}
This approximation is reasonable when there exists a unique optimal world model. As we will show, for complex target functions all members of the ensemble do not converge to a single model and as a result the white noise problem persists. A key limitation of this method is that memory usage grow linearly with size of the model ensemble. Disagreement-based curiosity is known as query by committee sampling \citep{seung1992query} in active learning. \\

\noindent \textbf{Novelty} \citep{bellemare2016unifying, dinh2016density, burda2018exploration} methods reward transitions with a low visitation count $\N(s, a, s')$. The prototypical novelty reward is: 
\begin{equation}
r(\bar{\rvs}, \rva, \bar{\rvs}') = \N(s_{t}, a_{t})^{-1/2}
\end{equation}
\cite{bellemare2016unifying} generalize visitation counts to pseudocounts for use in continuous state, action spaces. Novelty is a good surrogate reward when one seeks to maximize coverage over the transition space regardless of the learnability of the transition. This characteristic makes novelty reward prefer noisy data drawn from a high entropy distribution. Novelty reward is not adapted to the world model and thus has a propensity to be inefficient at reducing world model loss. \\

\noindent \textbf{Progress} \citep{schmidhuber_formaltheoryoffun, achiam2017surprise, graves2017automated} The key idea is to simply approximate the expectation involving $\mu$ in eq.~\ref{eq:cost} with the prediction gain on the history.
\begin{align}
r(\bar{\rvs}, \rva, \bar{\rvs}') = \LL_{H'}(\theta) - \LL_{H'}(\theta')
\label{eq:cost_emp}
\end{align}
where $H'$ is the augmented history after adding $(\bar{\rvs}, \rva, \bar{\rvs}')$.  
There is no guarantee the optimal policy with respect to eq. \ref{eq:cost_emp} is also an optimal policy with respect to eq. \ref{eq:cost} for every choice of $\mu$. However, we expect this history-based approximation of prediction gain to generate a data distribution that will be suitable for a wide array of $\mu$. If we think of the target $\omega$ as having easy, hard but doable, and impossible instances $(\rvx, \rvy)$, we expect such an agent to spend some time sampling easy, a good deal of time sampling the hard but doable, and little time on the impossible. For $\mu$ with support on easy data, little sampling is needed; for support on hard but doable, the greater proportion of samples is useful; and support on the impossible does not contribute to eq.~\ref{eq:cost} in the first place. 
Intuitively (if not formally), the progress curiosity approach should thus yield a data distribution\footnote{Actually, there is no guarantee that the stochastic process described by eq. \ref{eq:cost_emp} will converge in distribution at all, but the occupancy measure distribution will exist and the intuition presented here can be applied to that.} that is proportionate to the intrinsic learnability of the target $\omega$. 

Unlike eq. \ref{eq:cost}, policies based on eq. \ref{eq:cost_emp} are subject to sampling and robustness tradeoffs.  
New data gathered after parameter update from $\theta$ to $\theta'$ is expected to generate the most useful information for distinguishing $\LL_{H'}(\theta)$ from $\LL_{H'}(\theta')$. The fewer such samples there are in the history, the less well the empirical difference approximates true progress.
It is thus useful to compute the empirical difference in eq. \ref{eq:cost_emp} and perform the next parameter update after multiple samples from $\theta'$ have been gathered. On the other hand, the number of new samples between $\theta$ and $\theta'$ cannot be allowed to be too large, since less frequent progress updates would slow down the improvement of the dataset and thus, presumably, limit the efficiency of world model improvement. 
A natural compromise is to smooth the empirical progress measurement over multiple consecutive parameter changes, pooling datapoints to better approximate progress. 
This comes at the cost, however, of requiring access to model parameters at those multiple timepoints and some method for combining progress trajectory information as the model itself changes.  
All this is further complicated by the fact that the number of gradient descent steps taken when computing $\theta'$ from $\theta$ will itself be limited for computational reasons, making even the empirical progress computations noisier. 
Ensuring reliable and efficient approximation of progress thus requires careful choices of how often to update $\theta'$ and how to integrate information across multiple updates.  

$\delta$-\textbf{Progress}. One approach to such choices is given by $\delta$-progress \citep{achiam2017surprise, graves2017automated}, measures how much better the current ``new'' model $\theta_{new}$ is compared to an old model $\theta_{old}$, which, for forward prediction AWML, is
\begin{equation}
r(\bar{\rvs}, \rva, \bar{\rvs}') = \log{\frac{\omega_{\theta'}(\rvs' | \rvs, \rva)}{\omega_{\theta}(\rvs' | \rvs, \rva)}}
\simeq \log{\frac{\omega_{\theta_{new}}(\rvs' | \rvs, \rva)}{\omega_{\theta_{old}}(\rvs' | \rvs, \rva)}}.
\label{eq:progress_curr}
\end{equation}
Recall that $\mu$ is ideally a distribution whose support is learnable data with respect to model class $\Theta$. There are two steps of approximation in eq. \ref{eq:progress_curr}. The first step assumes that training on a sample $(\rvs, \rva, \rvs')$ affects the total validation loss on learnable data $\mu$ only through the reduction in loss on that particular sample. The second step assumes that future prediction gain is close to past prediction gain measured with respect to $\theta_{new}, \theta_{old}$. The choice of $\theta_{new}, \theta_{old}$ is crucial to the efficacy of the progress reward. A popular approach \citep{achiam2017constrained, graves2017automated} is to choose
\begin{equation}
\theta_{new} = \theta_{k}, \quad \theta_{old} = \theta_{k - \delta}, \quad \delta >0
\label{eq:kstep_progress}
\end{equation}
where $\theta_{k}$ is the model parameter after $k$ update steps using $P_{\ell}$. Intuitively, if the progress horizon $\delta$ is too large, we obtain an overly optimistic approximation of future progress. However if $\delta$ is too small, the agent may prematurely give up on learning hard transitions, e.g. where the next state distribution is very sharp. In practice, tuning the value $\delta$ presents a major challenge. Furthermore, the widely pointed out \citep{pathak2019disagreement} limitation of $\delta$-Progress is that the memory usage grows $\O(\delta)$, i.e one must store $\delta$ world model parameters $\theta_{k - \delta}, ..., \theta$. As a result it is intractable in practice to use $\delta > 3$ with deep neural net models. 

$\gamma$-\textbf{Progress}. Here we propose $\gamma$-Progress, the following choice of $\theta_{new}, \theta_{old}$ to overcome both hurdles faces by $\delta$-progress:
\begin{equation}
\boxed{
\theta_{new} = \theta, \quad \theta_{old} = (1 - \gamma) \sum_{i = 1}^{k - 1} \gamma^{k - 1 - i} \theta_{i}}
\label{eq:mix_progress}
\end{equation}
In words, the old model is a weighted mixture of past models where the weights are exponentially decayed into the past. $\gamma$-Progress can be interpreted as a noise averaged progress signal. Conveniently, $\gamma$-Progress can be implemented with a simple $\theta_{old}$ update rule:
\begin{equation}
\boxed{\theta_{old} \leftarrow \gamma \theta_{old} + (1 - \gamma) \theta_{new}}
\label{eq:mix_update_rule}
\end{equation}
Similar to eq.~\ref{eq:kstep_progress}, we may control the level of optimism towards expected future loss reduction by controlling the progress horizon $\gamma$, i.e a higher $\gamma$ corresponds to a more optimistic approximation. $\gamma$-Progress has key practical advantages over $\delta$-Progress: $\gamma$ is far easier to tune than $\delta$, e.g. we use a single value of $\gamma$ throughout all experiments, and memory usage is constant with respect to $\gamma$. Crucially, the second advantage enables us to tune the progress horizon so that the model does not prematurely give up on exploring hard transitions. The significance of these practical advantages will become apparent from our experiments.

\begin{figure*}[t!]
    \centering
    \includegraphics[width=\textwidth]{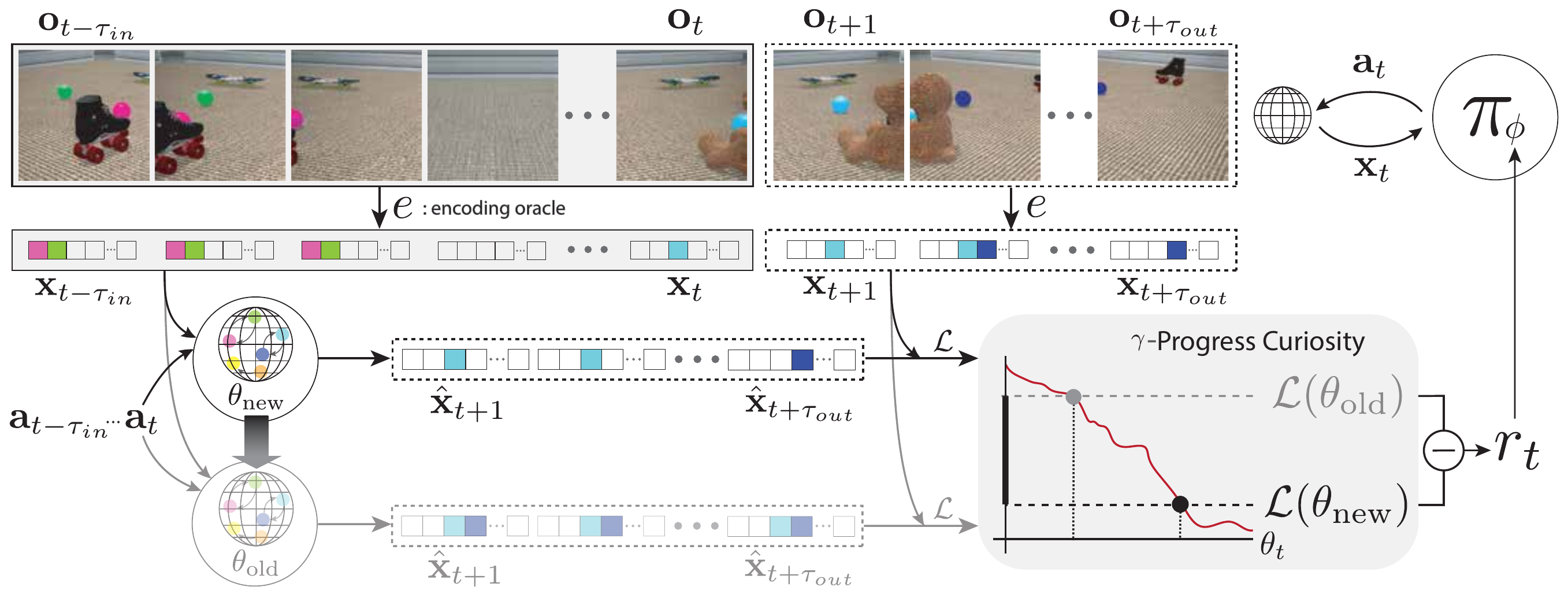}
    \vspace{-20pt}
    \caption{\textbf{Active World Model Learning with $\gamma$-Progress} The curious agent consists of a \emph{world model} and a \emph{progress-driven controller}. The curious agent's observations $\rvo_t$ are passed through an encoding oracle $e$ that returns an object-oriented representation $\rvx_t$ containing the positions of external agents that are in view, auxiliary object positions, and the curious agent's orientation. Both the new (opaque) and old (translucent) models take as input $\rvx_{t-\tau_{in}:t}$ and predict $\hat{\rvx}_{t:t+\tau_{out}}$. The old model weights, $\theta_{old}$, are slowly updated to the new model weights $\theta_{new}$. The controller, $\pi_{\phi}$, is optimized to maximize $\gamma$-Progress reward: the difference $\LL(\theta_{old}) - \LL(\theta_{new})$.}
    \label{fig:model}
\end{figure*}

\section{Methods}
\label{sec:methods}
In this section we describe practical instantiations of the two components in our AWML system: a {\em world model} which fits the forward dynamics and a {\em controller} which chooses actions to maximize $\gamma$-Progress reward. See Appendix \ref{sec:app:training_details} for full details on architectures and training procedures.  

\textbf{World Model} As the focus of this work is not to resolve the difficulty of representation learning from high-dimensional visual inputs, we assume that the agent has access to an oracle encoder $e: \O \rightarrow \X$ that maps an image observation $\rvo_t \in \O$ to a disentangled object-oriented feature vector $\rvx_t = (\rvx_{t}^{ext}, \rvx_{t}^{aux}, \rvx_t^{ego})$ where $\rvx_{t}^{ext} = (\tilde{\rvc}_{t}, \rvm_{t}) = (\tilde{\rvc}_{t, 1}, \ldots \tilde{\rvc}_{t, n_{ext}}, \rvm_{t, 1}, \ldots ,\rvm_{t, n_{ext}})$ contains information about the external agents; namely the observability masks $\rvm_{t, i}$ ($\rvm_{t, i} = 1$ if external agent $i$ is in curious agent's view at time $t$, else $\rvm_{t, i} = 0$) and masked position coordinates $\tilde{\rvc}_{t, i} = \rvc_{t, i}$ if $\rvm_{t, i} = 1$ and else $\tilde{\rvc}_{t, i} = \hat{\rvc}_{t, i}$. Here, $\rvc_{t, i}$ is the true global coordinate of external agent $i$ and $\hat{\rvc}_{t, i}$ is the model's predicted coordinate of external agent $i$ where $i = 1, \ldots , n_{ext}$. Note that the partial observability of the environment is preserved under the oracle encoder since it provides coordinates only for external agents in view. $\rvx_{t}^{aux}$ contains coordinates of auxiliary objects, and $\rvx_t^{ego}$ contains the ego-centric orientation of the curious agent. 

Our world model $\omega_{\theta}$ is an ensemble of component networks $\{\omega_{\theta^{k}}\}_{k = 1}^{N_{\mathrm{cc}}}$ where each $\omega_{\theta^{k}}$ independently predicts the forward dynamics for a subset $I_{k} \subseteq \{1, ..., \text{dim}(\rvx^{ext})\}$ of the input dimensions of $\rvx^{ext}$ corresponding to a minimal behaviorally interdependent group. For example, $\rvx^{ext}_{t:t+\tau, I_{k}}$ may correspond to the masked coordinates and observability masks of the chaser and runner external agents for times $t, t+1, ..., t+\tau$. We found that such a "disentangled" architecture outperforms a simple entangled architecture (see Discussion and Fig. \ref{fig:offline_asymptotic_performance}). We assume $\{I_{k}\}_{k = 1}^{N_{cc}}$ is given as prior knowledge but future work may integrate dependency graph learning into our pipeline. A component network $\omega_{\theta^{k}}$ takes as input $(\rvx^{ext}_{t - \tau_{in} : t, I_{k}}~,~\rvx^{aux}_{t - \tau_{in} : t}~,~\rvx^{ego}_{t - \tau_{in} : t}~,~\rva_{t - \tau_{in} : t + \tau_{out}})$, where $\rva$ denotes the curious agent's actions, and outputs $\hat{\rvx}^{ext}_{t : t + \tau_{out}, I_{k}}$. The outputs of the component network are concatenated to get the final output $\hat{\rvx}^{ext}_{t : t + \tau_{out}} = (\hat{\rvc}_{t:t+\tau_{out}}, \hat{\rvm}_{t:t+\tau_{out}})$. The world model loss is:  
\begin{align*}
    \LL(\theta, \rvx_{t-\tau_{in}:t+\tau_{out}}, \rva_{t-\tau_{in}:t+\tau_{out}}) = \sum_{t' = t}^{t + \tau_{out}} \sum_{i = 1}^{N_{ext}} \rvm_{t', i} \cdot \|\hat{\rvc}_{t', i} - \tilde{\rvc}_{t', i}\|_2 + \LL_{ce}(\hat \rvm_{t', i}, \rvm_{t', i})
\end{align*}
where $\mathcal{L}_{ce}$ is cross-entropy loss. We parameterize each component network $\omega_{\theta^{k}}$ with a two-layer Long Short-Term Memory (LSTM) network followed by two-layer Multi Layer Perceptron (MLP). The number of hidden units are adapted to the number of external agents being modeled. 

\textbf{The Progress-driven Controller} Our controller $\pi_{\phi}$ is a two-layer fully-connected network with 512 hidden units that takes as input $\rvx_{t-2:t}$ and outputs estimated Q-values for 9 possible actions which rotate the curious agent at different velocities. $\pi_{\phi}$ is updated with the DQN \citep{mnih2013playing} learning algorithm using the cost:
\begin{align}
c(\rvx_{t}) = \LL(\theta_{new}, \rvx_{t - \tau_{in} - \tau_{out}:t}, \rva_{t - \tau_{in} - \tau_{out}:t}) - \LL(\theta_{old}, \rvx_{t - \tau_{in} - \tau_{out}:t}, \rva_{t - \tau_{in} - \tau_{out}:t})
\end{align}
with $\gamma = 0.9995$ across all experiments. 

\begin{algorithm}[t]
\caption{AWML with $\gamma$-Progress} \label{prog_cur}
\textbf{Require:} progress horizon $\gamma$, step sizes $\eta_{\omega}, \eta_{Q}$ \\ 
Initialize $\theta_{new}, \phi$\\
\For{$k = 1, 2, ...$}{
    Update policy: $\pi_{\phi} \leftarrow \epsilon\text{-}greedy(Q_{\phi}, \epsilon - 0.0001)$ \\
    Sample $(\rvx, \rva, c) \sim \pi_{\phi}$ and place in Buffer $\mc{B}$ \\
    where $c = \LL(\theta_{new}, \rvx, \rva) - \LL(\theta_{old}, \rvx, \rva)$ \\
    \For{$j = 1, ..., M$}{ 
        Sample batch $b_{j} \sim \mc{B}$ 
        
        Update new world model: 
        $\theta_{new} \leftarrow \theta_{new} - \text{ADAM}(\theta_{new}, b_{j}, \eta_{\omega}, \LL)$

        Update old world model: 
        $\theta_{old} \leftarrow \gamma \theta_{old} + (1 - \gamma) \theta_{new}$ 

        Update Q-network with DQN \citep{mnih2015human}: 
        $\phi \leftarrow \text{DQN}(\phi, b_{j}, \eta_{Q})$
    }
}
\label{alg:awml}
\end{algorithm}

\section{Experiments}
\label{sec:experiments}

We evaluate the AWML performance of $\gamma$-Progress on two metrics: \emph{end performance} and \emph{sample complexity}. End performance is the inverse of the the final world loss after a larger number of environment interactions, and intuitively measures the ``consistency'' of the proxy reward with respect to the true reward. Sample complexity measures the rate of reduction in world model loss $\LL_{\mu}(\theta)$ with respect to the number of environment interactions. The samples from the validation distribution $\mu$ correspond to core validation cases we crafted for each behavior. On the reaching behaviors, for example, we validate the world model loss with objects spawned at new locations. For details on each behavior-specific validation case and metric computation, we refer readers to Appendix \ref{sec:app:validation}.

Experiments are run in two virtual worlds: Mixture and Noise world. In the Mixture world, the virtual environment is instantiated with external agents spanning four representative types: static, periodic, noise, and animate. This set up is a natural distillation of a real-world environment containing a wide spectrum of behaviors. In the Noise world, the environment is instantiated with three noise agents and one animate agent. This world stress-tests the noise robustness of $\gamma$-Progress. For each world, we run separate experiments in which the animate external agents are varied amongst the deterministic and stochastic versions of reaching, chasing, peekaboo, and mimicry agents (see Section \ref{sec:environment}). 

We compare the AWML performance of the following methods: 

\begin{figure}[H]
    \centering
    \includegraphics[width=\textwidth]{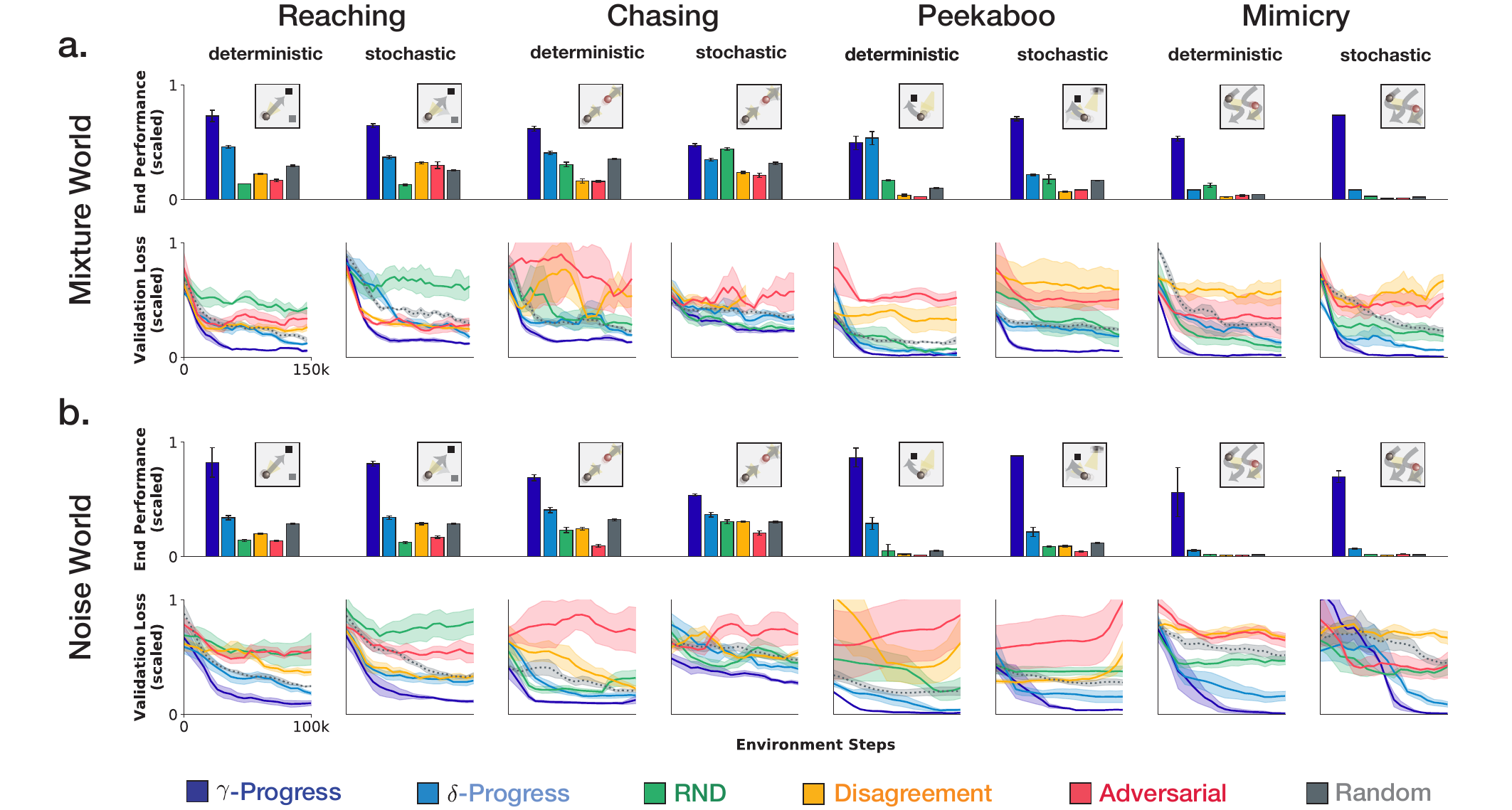}
    \vspace{-10pt}
    \caption{\textbf{AWML Performance and Sample Complexity}. The animate external agent is varied across experiments according to the column labels. End performance is the mean of the last five validation losses. Sample complexity plots show validation losses every 5000 environment steps. Error bars/regions are standard errors of the best $5$ seeds out of $10$. (a). \emph{Mixture World}: $\gamma$-Progress achieves lower sample complexity than all baselines on $7/8$ behaviors. Notably, $\gamma$-Progress also outperforms all baselines in end performance on $6/8$ behaviors. (b). \emph{Noise World}: $\gamma$-Progress is more robust to white noise than baselines and achieves lower sample complexity and higher end performance on $8/8$ behaviors. Baselines frequently perform worse than random due to noise fixation}
    \label{fig:sample_complexity}
\end{figure}

\vspace{-10pt}
\begin{SCtable}[1.175][h]
  \caption{\textbf{AWML Performance Summary.} Mean ratio of baseline end performance over Random baseline end performance (standard error in parentheses)}
  \label{table:aggregate_end_performance}
  \centering
  \begin{footnotesize}
  \begin{tabular}{lcc}
    \toprule
    & Mixture World & Noise World \\
    \midrule
    $\gamma$-Progress & \textbf{7.83} (3.57) & \textbf{13.79} (5.29) \\
    $\delta$-Progress & 2.2 (0.51) & 2.46 (0.55) \\
    RND & 1.25 (0.25) & 0.85 (0.10) \\
    Disagreement & 0.62 (0.10) & 0.76 (0.06) \\
    Adversarial & 0.62 (0.09) & 0.59 (0.10) \\
    \bottomrule
  \end{tabular}
  \end{footnotesize}
\end{SCtable}

\begin{itemize}
\item \textbf{$\gamma$-Progress (Ours)} is our proposed variant of progress curiosity which chooses $\theta_{old}$ to be a geometric mixture of all past models as in Eq. \ref{eq:mix_progress}. 

\item \textbf{$\delta$-Progress \citep{achiam2017surprise, graves2017automated}} is the $\delta$-step learning progress reward from Eq. \ref{eq:kstep_progress} with $\delta = 1$. We found that any $\delta > 3$ is impractical due to memory constraints.  

\item \textbf{RND \citep{burda2018exploration}} is a novelty-based method that trains a predictor neural net to match the outputs of a random state encoder. States for which the predictor networks fails to match the random encoder are deemed ``novel'', and thus receive high reward. 

\item \textbf{Disagreement \citep{pathak2019disagreement}} is the disagreement based method from Eq. \ref{eq:disagreement} with $N = 3$ ensemble models. We found that $N > 3$ is impractical due to memory constraints.  

\item  \textbf{Adversarial \citep{stadie2015predictive, pathak2017inverse}} is the prediction error based method from Eq. \ref{eq:adversarial}. We use the $\ell_{2}$ prediction loss of the world model as the reward. 

\item \textbf{Random} chooses actions uniformly at random among the 9 possible rotations. 
\end{itemize}

\subsection{AWML Performance.}
Fig.~\ref{fig:sample_complexity}a shows end performance (first row) and sample complexity (second row) in the Mixture world, and Fig.~\ref{fig:sample_complexity}b shows the same for the Noise World.
In the Mixture world, we see that $\gamma$-Progress has lower sample complexity than $\delta$-Progress, Disagreement, Adversarial, and Random baselines on all $8/8$ behaviors and outperforms RND on $7/8$ behaviors while tying on stochastic chasing. In the Noise world, we see that $\gamma$-Progress has lower sample complexity than all baselines on all $8/8$ behaviors. See Table~\ref{table:aggregate_end_performance} for aggregate end performance, and \textcolor{blue}{\url{https://bit.ly/31vg7v1}} for visualizations of model predictions.

\subsection{Attention control analysis}

\begin{figure}[t]
    \centering
    \includegraphics[width=\textwidth]{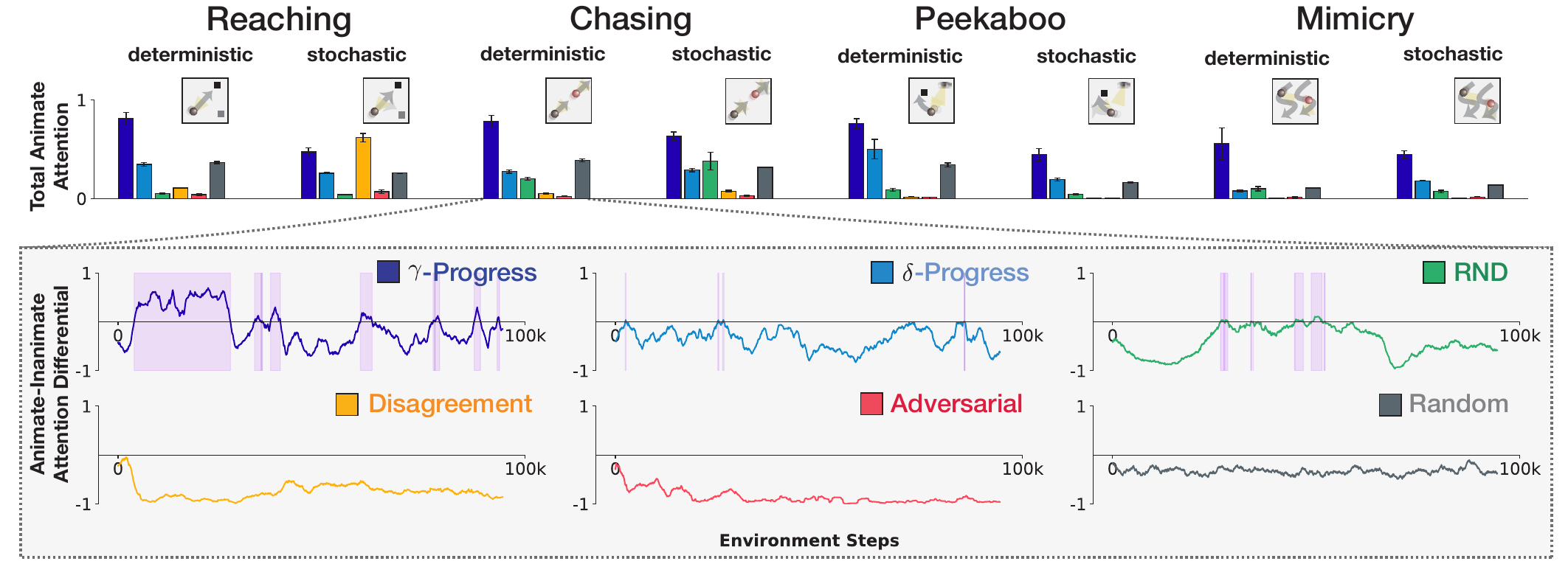}
    \vspace{-15pt}
    \caption{\textbf{Attention Patterns}. The bar plot shows the total animate attention, which is the ratio between the number of time steps an animate external agent was visible to the curious agent, and the time steps a noise external agent was visible. The time series plots in the zoom-in box show the differences between mean attention to the animate external agents and the mean of attention to the other agents in a 500 step window, with periods of animate preference highlighted in purple. Results are averaged across 5 runs. $\gamma$-Progress displays strong animate attention while baselines are either indifferent, e.g $\delta$-Progress, or fixating on white noise, e.g Adversarial.} 
    \label{fig:attention}
\end{figure}

Figure~\ref{fig:attention} shows the ratio of attention to animate vs other external agents for each behavior in the Mixture world as well as example animate-inanimate attention differential timeseries (for the Noise world, see Appendix~\ref{app:noise_world_plot}). The $\gamma$-Progress agents spend substantially more time attending to animate agents than do alternative policies. This increased animate-inanimate attention differential often corresponds to a characteristic attentional ``bump'' that occurs early as the $\gamma$-Progress curious agent focuses on animate external agents quickly before eventually ``losing interest'' as prediction accuracy is achieved. Strong animate attention emerges for $7/7$ behaviors when using $\gamma$-Progress. Please see appendix~\ref{app:further_attention_analysis} for a more in-depth analysis of how attention, and particular early attention, predicts performance and how curiosity signal predicts attention. 

Baselines display two distinct modes that lead to lower performance (Table~\ref{table:failure_modes}). The first is \textit{attentional indifference}, in which the curious agent finds no particular external agent interesting --- more precisely, we say that a curiosity signal choice displays attentional indifference if its average animate/inanimate ratio in the Mixture world is within two standard deviations of the Random policy's. $\delta$-Progress frequently had attentional indifference as the new and old world model, separated by a fixed time difference, were often too similar to generate a useful curiosity signal.

\begin{SCtable}[1.2]
    \caption{
    \textbf{Failure modes} Fraction of indifference and white noise failures, out of eight external agent behaviors.
    }
    \label{table:failure_modes}
    \centering
    \begin{footnotesize}
    \begin{tabular}{lcc}
    \toprule
    & Indifference & Noise Fixation \\
    \midrule
    $\gamma$-Progress & \textbf{0/8} & \textbf{0/8} \\
    $\delta$-Progress & 7/8 & 0/8 \\
    RND & 2/8 & 4/8 \\
    Disagreement & 0/8 & 7/8 \\
    Adversarial & 0/8 & 8/8 \\
    Random & 8/8 & 0/8 \\
    \bottomrule
    \end{tabular}
    \end{footnotesize}
\end{SCtable}

The second failure mode is \textit{white noise fixation}, where the observer is captivated by the noise external agents --- more precisely, we say that a curiosity signal choice displays white noise fixation if its average animate/inanimate ratio in the Noise world is more than two standard deviations below the Random policy's. RND suffers from white noise fixation due to the fact that our noise behaviors have the most diffuse visited state distribution. We also observe that for noise behaviors, a world model ensemble does not collectively converge to a single mean prediction, and as a result Disagreement finds the noise behavior highly interesting. Finally, the Adversarial baseline fails since noise behaviors yield the highest prediction errors. The white noise failure mode is particularly detrimental to sample complexity for  RND, Disagreement, and Adversarial, as evidenced by their below-Random performance in the Noise world.

\section{Discussion and Future Directions}
\label{sec:future}

In this work, we propose an Active World Model Learning agent that observes and interacts with an agent-rich 3D environment. 
The AWML agent learns a predictive world model of this environment, combining an agent-centric disentangled world model with a curiosity-driven action policy. 
A main contribution of this work is introducing $\gamma$-Progress, a computationally-tractable approximate estimator of expected information gain.
$\gamma$-Progress is sensitive and robust enough to discover \emph{de novo} a simple form of animate attention without having to have this concept built in, ``realizing'' that animate agents are more interesting to focus on than inanimate alternatives across a variety of animate agent types and inanimate distractors. The curious neural agent equipped with $\gamma$-Progress is better able to allocate scarce attentional resources in the partially observable environment, and is thus substantially more effective at learning world models in our agent-rich environment. 


\textbf{More realistic embodiments, richer tasks and behaviors, and real-world input streams.} 
While our environment does capture some key features of proto-social interactions, it is lacking in several important ways.
First, our AWML agent only has gaze-driven interactions with the enviroment, and external agents have no complex effectors or physical features capable of expressing social cues. 
Extending our current work with a more realistic embodiment (including agents with full motility, articulated effectors, and gaze cue markers) is an important direction, especially because some types of important proto-social behaviors --- such as mutual gaze coordination, gaze following, and pointing --- can only be expressed using richer avatars. 
In this work we have only targeted improved external agent prediction as the success metric, but an important next step, enabled by improved embodiment, will be to extend to the case of imitation learning, where the observing agent not only learns about others but also about how to do things itself.  
Another limitation is that only the single central observer is implemented as an AWML agent, with all the external-agent behaviors limited to hard-coded routines.
We seek to investigate true multi-agent scenarios where all agents are running a curious policy (perhaps at different stages of learning). 

In the present work, we have chosen to avoid the complication of having the learned components of our AWML system work directly with visual inputs, so as to focus on the challenging policy learning problems.  
However, forcing our agents (both observer and the external) to use pixel-based visual inputs will be an important next step.  
We expect that filling in this gap will be a meaningful challenge, especially given the need to integrate a working memory system to handle partial observability (determining e.g. when an agent has gone out of view and identifying it when it returns). 
Once these improvements in embodiment, input realism, and behavioral richness are made, we will seek to deploy AWML on a real-world robotic platform. 

\begin{wrapfigure}{R}{0.5\textwidth}
\vspace{-10pt}
\begin{center}
    \includegraphics[width=.5    \textwidth]{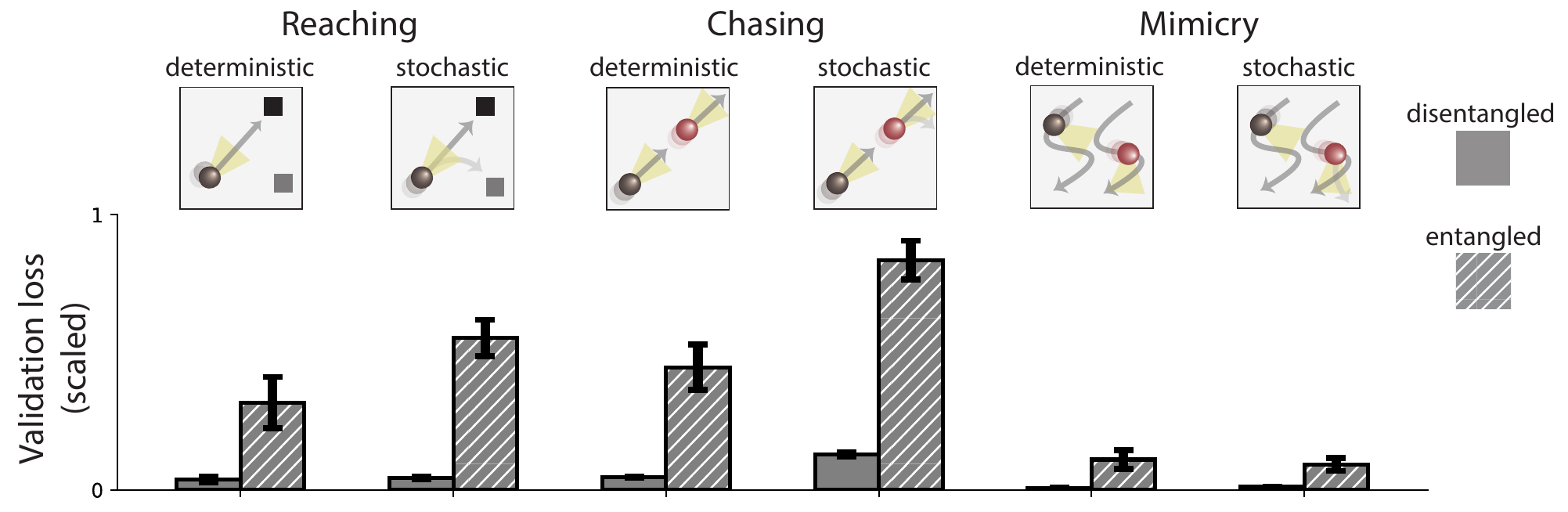}
\end{center}
\vspace{-10pt}
    \caption{\textbf{Asymptotic Model Performance} Final validation loss of the disentangled world model and entangled ablation on fixed dataset.}
    \label{fig:offline_asymptotic_performance}
\vspace{-10pt}
\end{wrapfigure}

\textbf{Disentangled world models and theory of mind.} 
To produce effective world models, we found it helpful to use an ``agent-centric'' disentangled architecture. 
However, is this choice necessary?  Do architectures not based on some form of agent-centric disentangling always fail to solve social multi-agent prediction problems? This is not obvious, since disentangling has proved a non-optimal (or at least non-necessary) strategy in some domains~\citep{locatello2018challenging, hong2016explicit}.  
As an initial investigation of this issue, we performed a pilot investigation of the effect of world model disentanglement on external-agent prediction performance (Fig. \ref{fig:offline_asymptotic_performance}).  
To ensure that this evaluation is independent of the choice of the policy controller (e.g. which type of curiosity, if any), our evaluation uses offline training datasets, one for each task in our current environment.\footnote[1]{Excluding peekaboo, since because the behavior is dependent on the observer's choices, no policy-independent offline training dataset can be constructed.} 
We compare performance between the agent-centric disentangled world model and an ``entangled'' or ``joint'' LSTM architecture that instead takes as input and predicts all external agents together. 
The disentangled architecture significantly outperforms the joint version: in fact, we originally began our investigation attempting to use the joint model, and only switched to the more complicated disentangled version when the former failed to work.  
Intuitively, the disentangled architecture performs better because it ignores spurious correlations between causally-unrelated events in the agent's data stream.
Formalizing this intuition mathematically and explaining why it may be particularly relevant in our current environment, in contrast to some other situations \citep{locatello2018challenging, hong2016explicit}, is an important future direction.
If it turns out that an agent-centric disentangled architecture is indeed robustly necessary, a natural question that will need to be resolved is how the interaction graph describing agent-agent dependencies can be estimated from observations, rather than known oracularly as here.  

Interestingly, this agent-centric disentangled architecture shares a key feature with the concept from cognitive science known as Theory of Mind (ToM).  
ToM describes the ability of one person to predict the behaviors of other people by inferring the others' mental states, such as beliefs, desires, and goals (\citep{astington1990developing, premack1978does, wellman1992child}).
A core, though often implicit, assumption of ToM is that separate predictive models are individually constructed for each non-self agent (or group of interacting non-self agents), and inferences about mental states are performed on a per-external-agent basis. 
Our disentangled model builds this as an inherent part of the architecture, and the performance improvements we observe from that choice loosely suggest that at least one possible function of ToM may be to enable statistical disentangling in highly partially-observed settings. 
Obviously, full ToM involves many other features beyond mere disentanglement that should enable effective external-agent model-making (e.g. model sharing and recursive representation of belief states) \citep{de2019common}, so the connection to the present work is at best partial.  
However, making this connection more concrete, and understanding how to build other key ToM features into an improved world model afford interesting avenues for future work. 

\begin{figure}
\begin{center}
    \includegraphics[width=\textwidth]{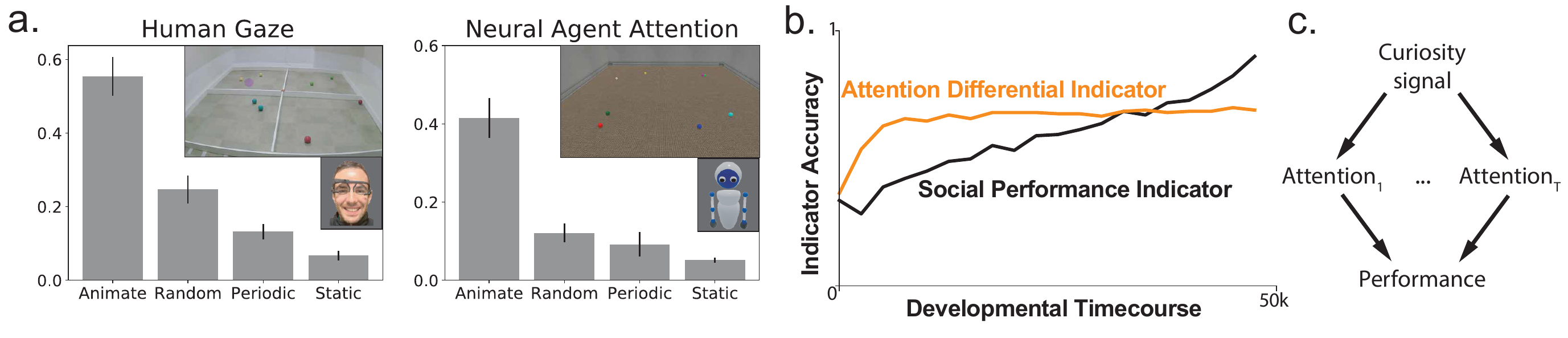}
\end{center}
\vspace{-20pt}
    \caption{\textbf{Modeling human behavior}. (a) Attentional preference in a pilot human behavior study and corresponding model preference. (b)  Accuracy of early indicators of final performance, as a function of time. Initially, the Attention Differential Indicator better predicts final performance, but as the time approaches final performance time, the Social Performance Indicator better predicts it. (c) Hypothesized factor diagram: curiosity signal determines attention, which determines final performance.}
    \label{fig:human_modeling_combined}
\vspace{-10pt}
\end{figure}

\noindent{\textbf{Toward quantitative models of human behavior and developmental variability.}}
Aside from AI uses, we hope that AWML might provide the basis of a quantitative model of human behavior. 
As a preliminary gesture in this direction, we have run a pilot human subject experiment (Fig. ~\ref{fig:human_modeling_combined}a) in which we exposed twelve adult human participants to static, periodic, animate, and noise stimuli using a display in which external agents were embodied with featurally-uniform spherical robots known as Spheros (\citep{kurkovsky2013sphero}).
Patterns of participants' attention were measured via a mobile eye tracker. 
We found that human adults display a clear animacy preference, quickly and reliably directing their gaze to focus on Spheros executing the more complex animate behaviors at the expense of predictable static/period or unpredictable random motion behaviors. 
They also exhibited a reliable pattern of relative attention across conditions.  
Comparing these measurements to those from our AWML agent, we found that the attention pattern generated by the $\gamma$-Progress policy is similar to that of the humans. 
While this pilot is too limited to draw any solid conclusion as to which model describes the human data best, it is illustrative of the type of comparisons we aim to make at much finer grain and greater scale in future work. 

Eventually, we would like to use the AWML agent as a model of intrinsically-motivated learning in early childhood.
Under this interpretation, the learning curves of the AWML agent should correspond to empirical developmental timecourses for the emergence of social attention in children. 
Improvements in external-agent prediction over time would be compared to empirical measures of changes in child social acuity, while attention allocation timecourses would be compared to how gaze patterns change developmentally. 

By extension, the AWML framework might also be used to describe the mechanisms underlying inter-subject variability in social development.
In this interpretation, variability in type of curiosity signal could represent a latent cause of developmental variability controlling both social acuity and attention allocation observables.
This connection is potentially plausible, since major causes of variability in social development, such as Autism Spectrum Disorder (ASD), are linked to differences in both low-level gaze preferences (\citep{jones2013attention}) and high-level social acuity (\citep{hus2014ados}). 
We hypothesize that this apparently wide and disparate spectrum of empirical variability might be accounted for by AWML-based computational model variants --- i.e. that ASD might in part be caused by systematic (and potentially genetically-linked~\citep{twinstudy}) differences between different children's mechanisms for intrinsically-motivated self-supervision. 

If correct, such ``computational etiology'' models could allow the design of model-driven diagnostics. 
To see what such a possibility might look like at a theoretical level, we consider the problem of how to predict, from early timepoint observations only, what the late (end-of-training) timepoint external-agent prediction performance of an AWML agent will be. 
Specifically, we estimate a statistical Attention Differential Indicator model, ATT$_{\leq T}$, which takes the agent's animate-inanimate attention differential (the quantity in Fig. \ref{fig:attention}a) up to time $T$ as input, and outputs predictions for the end-of-training performance (the quantity in Fig. \ref{fig:sample_complexity} barplots).  
The agent's curiosity policy is a latent source of variability that is hidden from the ATT$_{\leq T}$. 
As a baseline, we also trained a Social Performance Indicator model PERF$_{\leq T}$, which takes performance before time $T$ as input. 
As seen in Figure~\ref{fig:human_modeling_combined}b, ATT$_{\leq T}$ achieves reasonable prediction early in time, and throughout most of the ``developmental timecourse'' is actually a more accurate indicator of late performance than PERF$_{\leq T}$, the direct measurement of early-stage performance itself. 
The underlying reason why this occurs is possibly that attention differential is the very mechanism that eventually leads to better performance in better-performing models (e.g. $\gamma$-pogress), and (as seen in Fig. \ref{fig:attention}a) often manifests as an early ``bump'' in animate attention, allowing the ATT$_{\leq T}$ predictor to have high SNR. 
The overall structural equation model is conveyed by the factor diagram Figure~\ref{fig:human_modeling_combined}c --- for further details, see Appendix~\ref{sec:app:attention_diagnostic}. 

Currently, ASD diagnosis is done by expert clinicians, using observations of high-level behaviors (\citep{hus2014ados}). 
This method is subjective, expensive, and often too late --- the average diagnosis comes after 4 years of age, often preventing interventions during a critical period of development.
It would be of substantial utility if a simple and comparatively easy-to-estimate metric (such as gaze preference), measured early in development, could be used to predict social acuity in late childhood, which is highly salient for ASD outcomes.
Translating the above computational analysis into a real-world experimental population could lead to substantial improvements in diagnostics of developmental variability.

\section*{Acknowledgements}
This work was supported by the McDonnell Foundation (Understanding Human Cognition Award Grant No. 220020469), the Simons Foundation (Collaboration on the Global Brain Grant No. 543061), the Sloan Foundation (Fellowship FG-2018- 10963), the National Science Foundation (RI 1703161 and CAREER Award 1844724), the DARPA Machine Common Sense program, the IBM-Watson AI Lab, and hardware donation from the NVIDIA Corporation.

\bibliography{citations}

\newpage
\appendix 
\begin{center}
\Large \textbf{Appendix}
\end{center}


\section{Connections between General \& Conventional Active Learning}
\label{sec:app:gal_cal_connection}

\textbf{Pool-based Active Learning} is the same as Query Synthesis Active Learning with the only difference being $\A = \mc{D}_{pool}$ where $\mc{D}_{pool}$ is the initial pool of unlabelled data. 

\textbf{Stream Active Learning} is obtained by choosing $\S = \X \times \Y, \A = \{0, 1\}, P(\cdot | \rvs=(\rvx, \rvy), \rva) = \omega(\rvx)~\text{if}~\rva = 1~\text{else}~\delta(\rvy_{dum})$, and $c(\bar{\rvs} = (\rvs, H, \theta), \rva, \bar{\rvs}' = (\rvs', H', \theta')) = \LL_{val}(\theta) - \LL_{val}(\theta')$, where $\delta$ is the Dirac-delta function and $\rvy_{dum}$ is a dummy label that denotes the case when no label is returned by the oracle.



\section{Training Details}
\label{sec:app:training_details}

As shown in Algorithm \ref{alg:awml}, we interleave world model and policy updates while interacting with the environment. Specifically we update the both the world model and Q-network with 10 gradient steps per 40 environment steps. Both model updates begin after the buffer is filled with 1000 samples. 

\textbf{World Model}: 
We parameterize each component network $\omega_{\theta^{k}}$ with a two-layer Long Short-Term Memory (LSTM) network with 256 hidden units if $|I_{k}| = 1$ i.e., the causal group $k$ contains a single external agent, and 512 if $|I_{k}| \geq 2$ to ensure that the size of the parameter space scales with the input and output size. All networks are train using Adam with a learning rate of $1e\text{-}4$, $\beta_{1} = 0.9, \beta_{2} = 0.999$ and batch size $256$. 

The old model is synchronized with the new model weights once after $100$ world model updates. This "warm starts" the old model and prevents unreasonable large progress rewards at the start. We use a fixed value of the progress horizon $\gamma = 0.9995$ across all experiments. We found that any $0.9995 \leq \gamma \leq 0.9999$ attains similar results.

\textbf{Policy Learning}: 
For Q-network $Q_{\phi}$ updates we use the DQN algorithm
\citep{mnih2015human} with a discount factor of $\beta = 0.99$, a boostrapping horizon of $200$, a buffer size of $2e5$. Same as the world model, we train the Q-network using Adam with a learning rate of $1e\text{-}4$, $\beta_{1} = 0.9, \beta_{2} = 0.999$ and batch size $256$. The policy $\pi_{\phi}$ is an $\epsilon$-greedy exploration strategy with respect to $Q_{\phi}$. Specifically, $\epsilon$ is linearly decayed from $1.0$ to $0.025$ at a rate of $0.0001$ per environment step. 

\section{Validation Cases}
\label{sec:app:validation}

Here we describe validation protocol for each behavior. As data for the world model must be generated by interacting with the environment, what policy to use during validation is an important choice. As some behaviors are "interactive", i.e the external agent dynamics depend on the curious agent's actions, a naive policy that simply stares at the external agent may not elicit the core dynamics underlying the behavior. Thus, we hard-code the policy during validation to elicit the core dynamics for behavior and subsequently measure world model loss on the collected data. 

\textbf{Peekaboo}: The validation policy looks at the peekaboo external agent until it hides. The policy then keeps the peekaboo external agent in view so that when the agent "peeks" it immediately hides again. The validation loss measures the world model performance on predicting the dynamics of this peeking behavior which is representative of the core ``interactive'' nature of peekaboo. 

\textbf{Reaching}: At the start of validation, auxiliary objects are spawned at new locations which changes the trajectory of the reaching external agent. The validation policy then stares at the reaching external agent and validation loss is measured on the collected samples. This validation loss measures how well the world model has learned the contingency between the auxiliary object locations and the reaching external agent's movements. For example, a world model that has overfit to the external agent's trajectory for a particular set of auxiliary object locations will fail to generalize when auxiliary objects are spawned at new locations. 

\textbf{Chasing, Mimicry, Periodic, Static, Noise}: The validation policy simply stares at the external agents and validation loss is measured on the collected samples.

The validation losses shown in Figure \ref{fig:sample_complexity}a for the Mixture world is an average of the validation losses on the static, periodic, and animate external agents. The random agent is excluded from evaluation as there is virtually no learnable patterns in the behavior and averaging the large world model loss incurred on the random external agent could occlude the learning performance differences between curiosity signals on the other learnable external agents. For the Noise World, the shown validation losses in Figure \ref{fig:sample_complexity}b represent only the validation loss on the animate external agent.

\section{Noise World Attention}
\label{app:noise_world_plot}

\begin{figure}[h!]
    \centering
    \includegraphics[width=\textwidth]{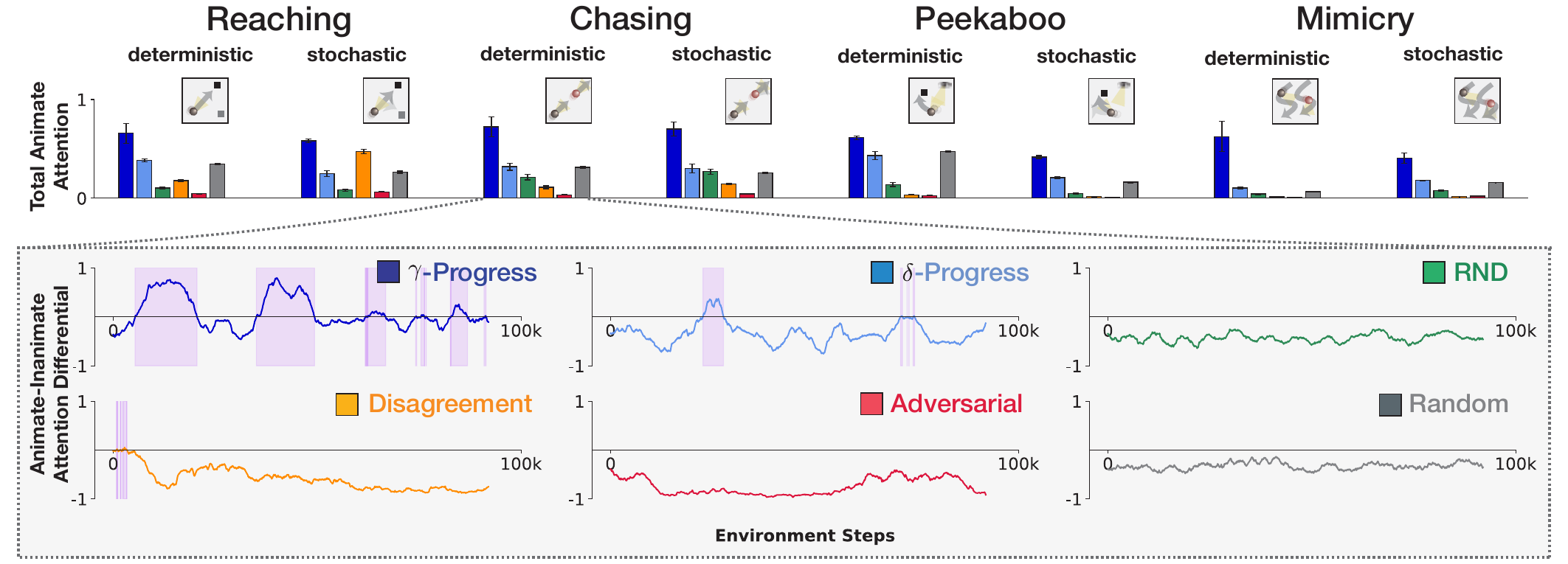}
    \vspace{-15pt}
    \caption{\textbf{Attention Patterns in Noise World}. The bar plot shows the total animate attention, which is the ratio between the number of time steps an animate external agent was visible and the number of time steps a noise external agent was visible. The zoom-in box plots show the differences between mean attention to the animate external agents and the mean of attention to the other agents in a 500 step window, with periods of animate preference highlighted in purple. Results are averaged across 5 runs. $\gamma$-Progress displays strong animate attention while baselines are either indifferent, e.g $\delta$-Progress, or fixating on white noise, e.g Adversarial.} 
    \label{fig:attention_noise}
    \vspace{-10pt}
\end{figure}


\section{Further attention analyses}
\label{app:further_attention_analysis}

Here we provide details of the early indicator analysis (Section~\ref{sec:future}) and a regression of what factors (curiosity signal, architecture, external agent behavior) best predict animate/inanimate attention ratios.

\subsection{Details of early indicator analysis}
\label{sec:app:attention_diagnostic}

We look to predict final performance $P_{\mathrm{final}}$ of a given agent, which we take to be the average of the final four validation runs. To make the modeling problem simple, we discretize this into a classification task by dividing validation performance into 3 equal-sized classes (``high'', ``medium'', and ``low'', computed separately for each external agent behavior), intuitively chosen to reflect performance around, at, and below that of random policy.

We consider two predictive models of final performance, one that takes as input early attention of the agent, and the other, early performance. Early performance may be quantified simply: given time $T$ (``diagnostic age'') during training, let $P_{\leq T}$ be the vector containing all validation losses measured up to time $T$. Early attention, however, is very high-dimensional, so we must make a dimensionality-reducing choice in order to tractably model with our modest sample size. Hence, we ``bucket'' average. Given choice of integer $B$, let
\begin{equation}
    A_{\leq T, B} = (\mathit{f}^{\mathrm{anim}}_{0 : \frac{T}{B}}, \mathit{f}^{\mathrm{rand}}_{ 0 :  \frac{T}{B}}, \mathit{f}^{\mathrm{anim}}_{\frac{T}{B}: \frac{2T}{B}}, \mathit{f}^{\mathrm{rand}}_{\frac{T}{B} : \frac{2T}{B}}, \ldots \mathit{f}^{\mathrm{anim}}_{\frac{(B - 1)T}{B} : T}, \mathit{f}^{\mathrm{rand}}_{\frac{(B - 1)T}{B} : T}),
\end{equation}
where $\mathit{f}^{\mathrm{anim}}_{a : b}$ and $\mathit{f}^{\mathrm{rand}}_{a : b}$ are the fraction of the time $t = a$ and $t = b$ spent looking at the animate external agent and random external agents respectively (so $A_{\leq T, B}$ is the attentional trajectory up to time $T$ discretized into $B$ buckets).

Finally, both models must have knowledge of the external agent behavior to which the agent is exposed --- we expect this to both have an effect on attention as well as the meaning of early performance and expected final performance as a result. Let $\chi_{\mathrm{BHR}}$ be the one-hot encoding of which external animate agent behavior is shown. 

We then consider models
\begin{enumerate}
    \item PERF$_{\leq T}$, which takes as input $P_{\leq T}$ and $\chi_{\mathrm{BHR}}$, and
    \item ATT$_{\leq T}$, which takes as input $A_{\leq T, B}$ and $\chi_{\mathrm{BHR}}$.
\end{enumerate}

Figure~\ref{fig:human_modeling_combined}b shows the plot of PERF$_{\leq T}$ and ATT$_{\leq T}$ accuracy as $T$ varies. We see that, up to a point, ATT$_{\leq T}$ makes a better predictor of final performance, and then PERF$_{\leq T}$ dominates. This confirms the intuition that attention patterns precede performance improvements. Intuitively, early attention predicts performance by being able to predict the sort of curiosity signal the agent is using, which predicts the full timecourse of attention (see~\ref{sec:app:attention_determinants}), which in turn predicts performance.

\subsection{Determinants of attention pattern}
\label{sec:app:attention_determinants}
To gain a finer-grained understanding of what, of the factors we vary (curiosity signal, world model architecture, and stimulus type) drives the attentional behavior of these active learning systems, we perform a linear regression. Specifically, we regress
\begin{equation} \label{eq:regression}
R_{\mathrm{animate}/\mathrm{noisy}} = a + b \cdot \chi_{\mathrm{CS}} + c \chi_{\mathrm{causal}} + d \cdot \chi_{\mathrm{BHR}} + \chi_{\mathrm{causal}} * e \cdot \chi_{\mathrm{IM}} + \epsilon
\end{equation}
Here $R_{\mathrm{animate}/\mathrm{noisy}}$ is the ratio of animate to noisy attention, $\chi_{\mathrm{CS}}$ is a one-hot encoding of curiosity signal (all zeros if random policy), $\chi_{\mathrm{causal}}$ is an indicator set to $1$ if the architecture is causal, $\chi_{\mathrm{BHR}}$ is a one-hot encoding of animate external agent behavior shown (all zeros if deterministic reaching), and $a,  b, c, d, e$ are fixed effects ($e$ measures an interaction effect).

Over $371$ individual active learning runs, an ordinary least squares regression achieves an adjusted $R^2$ of $.44$. Please see Table~\ref{table:regression} for details. We found that $\gamma$-Progress receives significant positive weight, while Disagreement and Adversarial receive significant negative weight, with the other curiosity signals having an effect close to that of random policy. In addition, we fail to find a significant effect due to architecture and most external agent behaviors, with two external agent behavior exceptions. In sum, we find that, of the architectural and curiosity signal variations we tested, curiosity signal strongly drives behavior whereas architecture plays an insignificant role.

\begin{table*}
\caption{
\textbf{Attention regression.} Regression model of animate/noisy attention, according to Equation~\ref{eq:regression}. Coefficient values found, and uncorrected p-value for 2-sided t-tests, with significance at the $.05$ level in bold.}
\vspace{-0.1cm}
\label{table:regression}
\vskip 0.15in
\begin{center}
\begin{small}
\begin{sc}
\begin{tabular}{lcc}
\toprule
Coefficient & value & P > |t| \\
\midrule
constant & .80 & .001 \\
$\gamma$-Progress & \textbf{2.24} & .000 \\
$\delta$-Progress & .08 & .788 \\
RND & -.53 & .064 \\
Disagreement & \textbf{-.70} & .014 \\
Adversarial & \textbf{-.79} & .006 \\
\midrule
Causal architecture & .014 & .959 \\
\midrule
stochastic reaching & .14 & .493\\
deterministic chasing & .25 & .222 \\
stochastic chasing & \textbf{.45} & .029 \\
deterministic peekaboo & -.08 & .682\\
stochastic peekaboo & .02 & .920\\
mimicry & \textbf{.56} & .006\\
\midrule
causal $\times \gamma$-Progress & -.32 & .408 \\
causal, $\times \delta$-Progress & .06 & .868 \\
causal $\times$ RND & .03 & .935 \\
causal $\times$ Disagreement & .23 & .555 \\
causal $\times$ Adversarial & -.09 & .813 \\
\bottomrule
\end{tabular}
\end{sc}
\end{small}
\end{center}
\end{table*}







\end{document}